\documentclass[letter,conference]{IEEEtran}
\IEEEoverridecommandlockouts
\usepackage{cite}
\usepackage{amsmath,amssymb,amsfonts}
\usepackage{algorithmic}
\usepackage{graphicx}
\usepackage{textcomp}
\usepackage{xcolor}
\usepackage[caption=false,font=footnotesize]{subfig}
\usepackage{overpic}
\usepackage{multirow}
\usepackage{xcolor}

\def\BibTeX{{\rm B\kern-.05em{\sc i\kern-.025em b}\kern-.08em
    T\kern-.1667em\lower.7ex\hbox{E}\kern-.125emX}}

\newcommand{\todo}[1]{}
\renewcommand{\todo}[1]{\textcolor{red}{@All: {#1}}}
\newcommand{\todoMF}[1]{}
\renewcommand{\todoMF}[1]{\textcolor{orange}{@MF: {#1}}}
\newcommand{\todoSG}[1]{}
\renewcommand{\todoSG}[1]{\textcolor{purple}{@SG: {#1}}}

\newcommand{\lorem}[1]{}
\renewcommand{\lorem}[1]{\todo{REMOVE}~\textcolor{gray}{Lorem ipsum dolor sit amet, consetetur sadipscing elitr, sed diam nonumy eirmod tempor invidunt ut labore et dolore magna aliquyam erat, sed diam voluptua. At vero eos et accusam et justo duo dolores et ea rebum. Stet clita kasd gubergren, no sea takimata sanctus est Lorem ipsum dolor sit amet. Lorem ipsum dolor sit amet, consetetur sadipscing elitr, sed diam nonumy eirmod tempor invidunt ut labore et dolore magna aliquyam erat, sed diam voluptua. At vero eos et accusam et justo duo dolores et ea rebum. Stet clita kasd gubergren, no sea takimata sanctus est Lorem ipsum dolor sit amet.
Lorem ipsum dolor sit amet, consetetur sadipscing elitr, sed diam nonumy eirmod tempor invidunt ut labore et dolore magna aliquyam erat, sed diam voluptua. At vero eos et accusam et justo duo dolores et ea rebum. Stet clita kasd gubergren, no sea takimata sanctus est Lorem ipsum dolor sit amet. Lorem ipsum dolor sit amet, consetetur sadipscing elitr, sed diam nonumy eirmod tempor invidunt ut labore et dolore magna aliquyam erat, sed diam voluptua. At vero eos et accusam et justo duo dolores et ea rebum. Stet clita kasd gubergren, no sea takimata sanctus est Lorem ipsum dolor sit amet.}}

\newcommand{\loremSmall}[1]{}
\renewcommand{\loremSmall}[1]{\todo{REMOVE}~\textcolor{gray}{Lorem ipsum dolor sit amet, consetetur sadipscing elitr, sed diam nonumy eirmod tempor invidunt ut labore et dolore magna aliquyam erat, sed diam voluptua. At vero eos et accusam et justo duo dolores et ea rebum. Stet clita kasd gubergren, no sea takimata sanctus est Lorem ipsum dolor sit amet. Lorem ipsum dolor sit amet, consetetur sadipscing elitr, sed diam nonumy eirmod tempor invidunt ut labore et dolore magna aliquyam erat, sed diam voluptua. At vero eos et accusam et justo duo dolores et ea rebum.}}

\newcommand{\name}{HPERL}
\newcommand{\basename}{RGB baseline}

\begin{document}

\title{\name: 3D Human Pose Estimation\\from RGB and LiDAR
}

\author{
\IEEEauthorblockN{
Michael Fürst\IEEEauthorrefmark{1}\IEEEauthorrefmark{2},
Shriya T. P. Gupta\IEEEauthorrefmark{1}\IEEEauthorrefmark{3},
René Schuster\IEEEauthorrefmark{2},
Oliver Wasenmüller\IEEEauthorrefmark{2}\IEEEauthorrefmark{4} and
Didier Stricker\IEEEauthorrefmark{2}\IEEEauthorrefmark{5}}
\IEEEauthorblockA{\IEEEauthorrefmark{1}Equal contribution}
\IEEEauthorblockA{\IEEEauthorrefmark{2}DFKI - German Research Center for Artificial Intelligence, \texttt{firstname.lastname@dfki.de}}
\IEEEauthorblockA{\IEEEauthorrefmark{3}Birla Institute of Technology and Science (BITS) Pilani, \texttt{f20160060@goa.bits-pilani.ac.in}}
\IEEEauthorblockA{\IEEEauthorrefmark{4}University of Applied Sciences Mannheim, \texttt{o.wasenmueller@hs-mannheim.de}}
\IEEEauthorblockA{\IEEEauthorrefmark{5}University of Kaiserslautern - TUK}
}

\maketitle

\begin{abstract}

In-the-wild human pose estimation has a huge potential for various fields, ranging from animation and action recognition to intention recognition and prediction for autonomous driving.
The current state-of-the-art is focused only on RGB and RGB-D approaches for predicting the 3D human pose.
However, not using precise LiDAR depth information limits the performance and leads to very inaccurate absolute pose estimation.
With LiDAR sensors becoming more affordable and common on robots and autonomous vehicle setups, we propose an end-to-end architecture using RGB and LiDAR to predict the absolute 3D human pose with unprecedented precision.
Additionally, we introduce a weakly-supervised approach to generate 3D predictions using 2D pose annotations from PedX~\cite{pedx}. 
This allows for many new opportunities in the field of 3D human pose estimation.
\end{abstract}


\section{Introduction} \label{sec:introduction}

Human pose estimation and understanding is the foundation for intention recognition and action recognition.
In the context of fully autonomous or highly automated vehicles, it is essential to recognize and understand the pointing gestures of a police officer or other traffic participants.
The overall body pose also enables the estimation of whether a pedestrian is looking at a vehicle and waiting or crossing the street without seeing the car. Thus, it allows the automated car to react even before the pedestrian is on the road.
Furthermore, it can help with the rotation ambiguity for pedestrians. While it is debatable if the foot, hip or torso direction is the front of a pedestrian, with human pose estimation there is no need for a decision, since all joints are provided and a more detailed understanding is enabled.

However, there is presently a lack of human pose estimation approaches for pedestrians.
Currently most approaches in human pose estimation focus on controlled environments, and the few that handle in-the-wild scenarios do not focus on the specific situation of pedestrian detection in autonomous driving.
Autonomous vehicles need a good detection rate.
Furthermore, algorithms should be tuned towards false positives rather than false negatives, since the latter puts the pedestrians in great danger.
In contrast to most datasets and algorithms focusing on human pose estimation, the distance at which pedestrian detection happens is a challenge.
Relevant pedestrians on the sidewalk are typically 5-50 meters away from the ego-vehicle.

\begin{figure}[!t]
	\begin{center}
	    \begin{overpic}[width=0.98\linewidth,,tics=10]{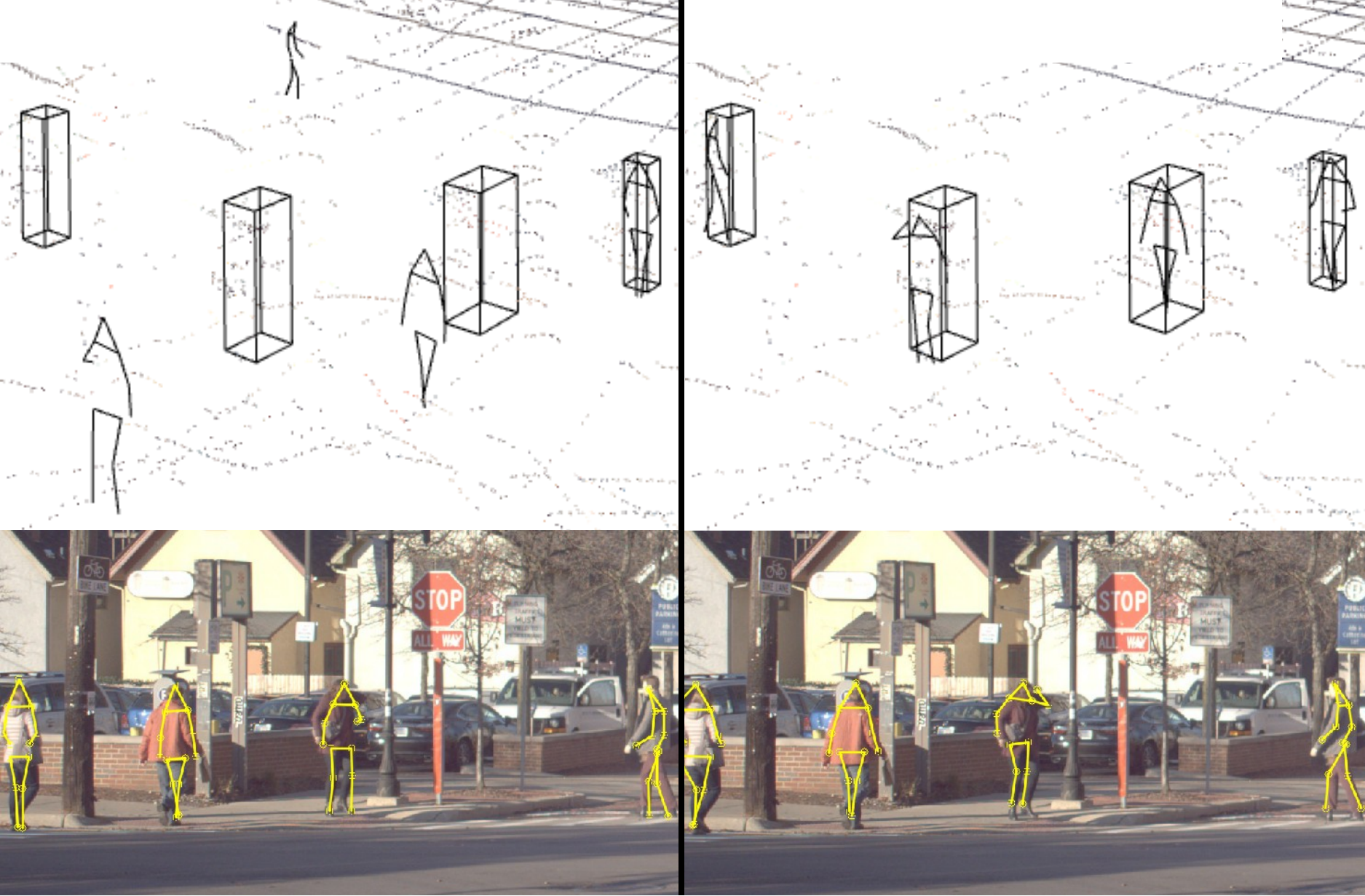}
	        \put(1,62){RGB only}
	        \put(51,62){\name~(RGB + LiDAR)}
		\end{overpic}
    \end{center}
    \caption{Depth ambiguity is solved by incorporating LiDAR information. The visualization of the predicted 3D poses and the ground truth 3D bounding boxes shows a poor performance for the RGB only case due to the depth ambiguity. But our \name~can precisely predict the poses and their absolute position, using LiDAR information. In 2D image space, the depth ambiguity leads to visually appealing results for both approaches.} 
    \label{fig:title}
\end{figure}

Moreover, with LiDAR sensors becoming more affordable and being used as a main sensor for other tasks in this field, there is the opportunity to not only rely on RGB as the current state-of-the-art does, but to use LiDAR as an additional input modality.
In 3D object detection, it has been shown that the addition of LiDAR enables a game changing precision.
We are the first to show similar insights for human pose estimation using our \name~(Fig.~\ref{fig:title}).

To make 3D human pose estimation precise enough for the demands of autonomous driving, we propose:
\begin{itemize}
    \item A novel end-to-end architecture for multi-person 3D pose estimation that fuses RGB images and LiDAR point clouds for superior precision,
    \item a weakly-supervised training procedure for simultaneous 2D and 3D pose estimation using only 2D pose labels,
    \item evaluation metrics to assess the 3D performance of our approach without expensive 3D pose annotations.
\end{itemize}

\section{Related Work} \label{sec:related_work}

Faster R-CNN\cite{fasterrcnn} is one of the most influential object detectors.
Inspiring many approaches, it is also at the core of our work.
It has a region proposal network, that predicts regions of interest in the image and then refines those predictions with a second stage.
Approaches following this scheme can be observed in many fields related to our work.
In the following sections, we briefly introduce all the associated fields.


\subsection{3D Detection proves importance of LiDAR}

For 3D object detection in the field of autonomous driving, there is a division of approaches based on the sensor modalities used for detection.
There are RGB only approaches, LiDAR only approaches and RGB+LiDAR approaches.
RGB only approaches are actively researched~\cite{direct3d} but cannot achieve the performance of LiDAR approaches~\cite{pointpillars, pointrcnn}. Most approaches~\cite{ipod, mv3d, VMVS, fpointnet} are in the RGB+LiDAR category, but majorly influential to our \name~are  AVOD~\cite{avod} and LRPD~\cite{fuerst2020lrpd}.
LRPD~\cite{fuerst2020lrpd} has shown that for detecting far away pedestrians precisely, the details of the RGB image and the precision of LiDAR are both essential. This indicates that RGB+LiDAR fusion can yield great performance improvements for precise human pose estimation.

AVOD~\cite{avod} follows a two stage approach like Faster R-CNN. In the first stage, they generate region proposals from the RGB and LiDAR inputs and fuse them using the RoI crops.
The second stage then operates on the RoI feature crops like the refinement stage of Faster R-CNN, with the main conceptual difference being that the regression is for 3D boxes instead of 2D boxes.
This structure allows it to be adapted to human pose estimation approaches following the Faster R-CNN~\cite{fasterrcnn} schema.

\begin{figure}[t]
    \begin{center}
	    \begin{overpic}[width=0.98\linewidth,,tics=10]{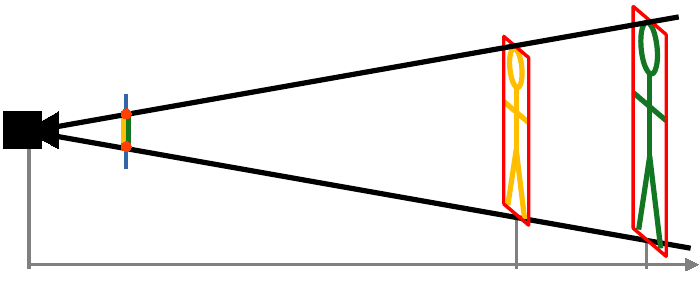}
	        \put(10,33){Image Plane}
		    \put(0.5,0){$d_{cam}=0$}
   		    \put(72,0){$d_1$}
   		    \put(90,0){$d_2$}
   		    \put(99,0){$d$}
		\end{overpic}
    \end{center}
    \caption{The two pedestrians (yellow, green) appear to be of the same size in the RGB image, even though they have different distances from the camera. A slight change in height can have an impact on the estimated distance. A network can still partially reconstruct the depth from other cues, but this is more difficult than with the correct LiDAR depth information.}
    \label{fig:sota}
\end{figure}
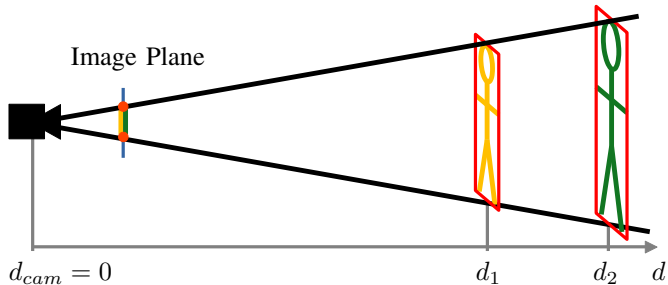

\subsection{2D Human Pose Estimation}

In the past, 2D human pose estimation has been successfully solved by various approaches on RGB images only.
DeepPose~\cite{toshev2014deeppose} applied CNNs in a cascaded regressor for 2D human pose estimation, whereas Tompson et al. \cite{tompson2015efficient} predicted heatmaps for the joints instead of direct regression.
In~\cite{wei2016convolutional} and~\cite{newell2016stacked}, the heatmap idea is further improved upon.
With the advent of multi-person pose estimation, two main categories of pose estimators emerged.

\subsubsection{Bottom-Up}

Approaches predicting a heatmap of the joint positions first, and then combining the joints into human poses are called bottom-up methods~\cite{pishchulin2016deepcut, cao2017realtime, sun2019deep}.

\subsubsection{Top-Down}

These follow the opposite approach, by first predicting a bounding box around the person and then regressing the joints of that person~\cite{fang2017rmpe, papandreou2017towards, huang2017coarse, chen2018cascaded}. As a direct descendant of Faster R-CNN~\cite{fasterrcnn}, Mask R-CNN~\cite{he2017mask} is the most adaptable approach from this category proving the strength of its architecture in bounding box regression, segmentation and human pose estimation.
DensePose~\cite{alp2018densepose} is a descendant of Mask R-CNN that maps the UV-coordinates of a 3D model to a person in the image, demonstrating the versatility of top-down estimators.

Our approach is inspired by Faster R-CNN~\cite{fasterrcnn} and can be attributed to the top-down category. This method was chosen, as 3D object detectors with fusion typically rely on Faster R-CNN like approaches.

\subsection{3D Human Pose Estimation}

Li et al.~\cite{li20143d} solve the 3D pose estimation task by directly regressing the joint positions and then detecting the actual 3D joints. In contrast, Chen et al. \cite{chen20173d} predict 2D poses, match them to a 3D pose library and use the best match as the 3D pose. Similarly, Martinez et al. \cite{martinez2017simple} use a simple neural network to predict 3D poses from the 2D poses. But Zhou et al.~\cite{zhou2017towards} observe that the sequential nature of separated sequential approaches~\cite{chen20173d, martinez2017simple} hinders performance. So, they integrate the learning process by having images from 2D in-the-wild and 3D indoor datasets in one batch. The 2D module is trained with 2D images and the 3D module is trained using 2D constraints and 3D regression data.

Further, there are RGB-D approaches like~\cite{zimmermann20183d, girshick2011efficient}. But as VNect~\cite{mehta2017vnect} shows, RGB-D methods suffer from limited application domains, mostly restricted to indoors. Moreover, the precision is not superior to RGB only methods.

LCR-Net~\cite{lcrnet, lcrnet++} is a simple yet effective representative of the 3D pose estimation category.
Its overall architecture is similar to Faster R-CNN~\cite{fasterrcnn}.
However, instead of just predicting regions of interest, it adds pose proposals, which are then refined in a second stage.
The refinement has multiple parallel regression heads, one for each pose proposal, allowing a high degree of specialization in the poses.

Although 3D object detection has shown the importance of LiDAR, mainly for resolving scale ambiguity errors as in Fig.~\ref{fig:sota}, none of the presented pose estimation approaches use a fusion of RGB and LiDAR.
Analysing the state-of-the-art, Faster R-CNN~\cite{fasterrcnn} style methods in 3D object detection (AVOD) and in 3D human pose estimation (LCR-Net) share a common structure that can be exploited.
To the best of our knowledge, there have been no experiments on the fusion of RGB and LiDAR for 3D human pose estimation.

\section{Approach} \label{sec:approach}
\begin{figure*}[t]
    \begin{center}
	    \begin{overpic}[width=0.98\linewidth,,tics=10]{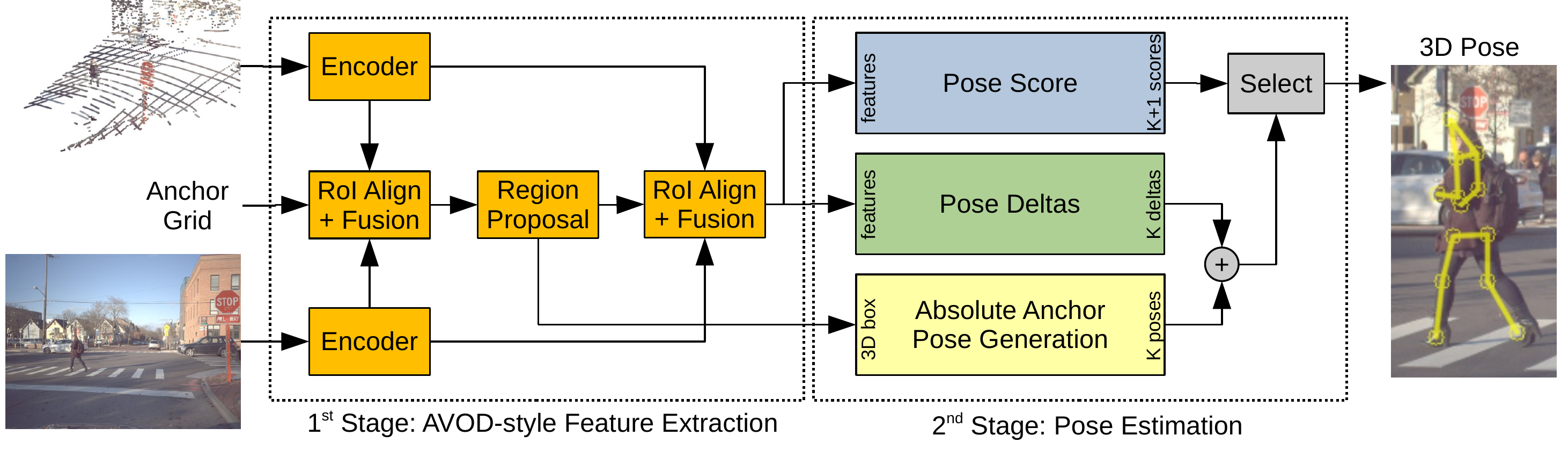}
		\end{overpic}
    \end{center}
    \caption{Our \name~architecture processes the RGB images and LiDAR point clouds as input modalities, using an RPN based on AVOD~\cite{avod} as the \textit{Feature Extraction Stage}. Inspired by LCR-Net~\cite{lcrnet}, our \textit{Pose Estimation Stage} predicts scores and deltas for the \textit{K} anchor poses. In contrast to other approaches, the anchor poses are generated from the 3D boxes of the first stage. By adding the deltas to these anchors and selecting based on the classification scores, the poses are predicted. These poses are then in a last step combined and filtered, whereas there may be multiple proposals per pedestrian.}
    \label{fig:architecture}
\end{figure*}
Here we outline the main components of our end-to-end trainable pose estimation network, with the first stage as the Region Proposal Network (RPN) and the second stage composed of the classification and regression branches (Fig.~\ref{fig:architecture}). We use an AVOD~\cite{avod} inspired first stage for \name~and a Faster R-CNN~\cite{fasterrcnn} inspired first stage for the \basename. As for the second stage, we use an LCRNet~\cite{lcrnet} inspired module in both cases. Thus, we perform the pose estimation in a top-down approach by first generating the region proposals and then estimating the human poses in the defined regions.

\subsection{Network Architecture}

\subsubsection{Multimodal Feature Extraction}
For the case of using both RGB and LiDAR data as input modalities, we first process the LiDAR point clouds by following the procedure in MV3D~\cite{mv3d} to create a six channel Bird's Eye View (BEV) representation.
The first stage of AVOD~\cite{avod} has two parallel VGG-16 modules for extracting features from the RGB and BEV inputs.
We modified these VGG-16 modules to use group normalization and 256 output channels in the feature maps.
Using the anchor grid defined in Section~\ref{Anchor Generation}, we project the 3D anchors onto the respective views and apply RoI align to crop the feature maps. The channel dimension is reduced to one by a fully connected layer, and the RGB and LiDAR views are averaged. Then, the objectness scores and regression offsets for the region proposals are predicted.

In contrast to AVOD~\cite{avod}, we use the RoI align operation to extract the features for a region proposal. RoI align avoids rounding off operations and preserves the spatial information, helping the overall performance of the network~\cite{maskrcnn}.
But unlike AVOD~\cite{avod}, the two streams of cropped RGB and LiDAR features are concatenated instead of averaged, preventing loss of information.
These features are then passed to the second stage of \name.

\subsubsection{Unimodal Feature Extraction}
For the baseline model having only RGB data as the input modality,
we use the first stage of Faster R-CNN~\cite{fasterrcnn} with a Resnet50~\cite{resnet50} feature extractor and a Feature Pyramid Network (FPN)~\cite{fpn} backbone. The weights are initialized from a COCO~\cite{coco} pre-trained version provided in the TorchVision library.
For this network, the RoI align operation is used to crop and resize the features to enable a fair comparison between the multimodal and unimodal approaches.


\subsubsection{Classification and Regression}
Based on the RoI features of the first stage, the second stage of our model classifies the proposals and predicts the regression deltas.
A fully connected layer is used for classifying each region proposal into one of the $K$ anchor poses or the background.
Another parallel fully connected layer predicts a set of $5\times J\times (K+1)$ pose deltas.
Here, $J=13$ is the number of joints, 5 represents the two values for 2D regression and three values for 3D regression. These pose deltas are then added to the anchor pose proposals to regress the actual 2D and 3D poses.


\subsection{Anchor Generation}
\label{Anchor Generation}
\begin{figure}[t]
    \centering
	\includegraphics[width=0.99\linewidth]{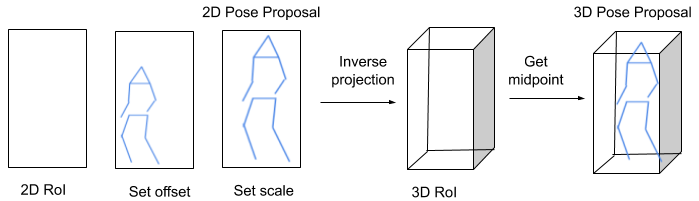}
    \caption{The pose proposals are generated by fitting the anchor poses into the predicted RoIs during inference. This is done by offsetting the anchor poses by an amount equal to the lowermost coordinates of the bounding box and then scaling them by the width and height of the RoI.}
    \label{fig:anchor_generation}
\end{figure}

\subsubsection{Anchor Boxes}
For the first stage of \name, we pass a pre-defined grid of 3D anchor boxes which is defined by the ground plane and area extents. The ground plane for our task is represented using the point-normal form of a plane $n \cdot (r - r_0) = 0$, with a normal $n = (a, b, c)$ and a point $r_0 = (x_0, y_0, z_0)$.
We define $n = (0, -1, 0)$ to match our known camera orientation.
The offset point $r_0$ is estimated using the RANSAC~\cite{ransac} algorithm with an additional offset of 1.8m to cover the ground.

We compute x and z ranges of the area extents by taking the minimum and maximum values of 3D pedestrian locations in the ground truth. The anchors are distributed over these area extents with a stride of 0.2m and the corresponding y coordinates are computed using the plane equation.

\subsubsection{Anchor Poses}
In order to choose a representative set of pedestrian poses, we define eight anchor poses which are a subset of the anchor poses used in LCRNet~\cite{lcrnet}. Amongst these, we exclude all the half body anchor poses because the pedestrian pose estimation task has only full body poses. Out of the remaining ones, we choose the ones that have a non-zero occurrence in the PedX~\cite{pedx} dataset. To align the anchor poses to the world coordinate system, we use the re-alignment procedure described in LCRNet~\cite{lcrnet}. In addition, we negate the y coordinates, as the negative y direction is the up-axis in our system. During the training phase, since there is no 3D ground truth available to assign the target deltas directly, we create the pose proposals as a pre-processing step using the ground truth bounding boxes. We add the predicted deltas to these pose proposals and train our model using only the 2D pose annotations and the projected 3D predictions. For inference, the pose proposals are generated by fitting anchor poses into the predicted RoIs as depicted in Fig.~\ref{fig:anchor_generation}.

\subsection{Loss Computation}
Since we aim to simultaneously predict the 2D and 3D poses in our model, we use a weighted multitask loss function composed of the RPN losses, the classification loss, the 2D loss and the projected 3D loss as follows:
\begin{equation}
L_{total} = L_{RPN} + L_{cls} + L_{2D} + L_{3D}
\end{equation}

\subsubsection{RPN Loss}
$L_{RPN}$ is composed of two components - the objectness loss $L_{obj}$ and the box regression loss $L_{reg}$. For \name, we compute these as specified in the first stage of AVOD~\cite{avod} using 3D ground truth boxes as the targets. Whereas for the \basename, we compute the $L_{RPN}$ as in Faster R-CNN~\cite{fasterrcnn} with the targets as 2D ground truth boxes.

\subsubsection{Anchor Pose Classification Loss}
Assignment of the anchor pose ground truth is a two step process.
First a categorization in foreground and background is done by IoU matching to the ground truth, then for foreground objects a similarity score is used to assign the best anchor pose.

The IoU computation between the ground truth and predicted RoIs varies with input modalities.
For the 3D RoIs of \name, we project them into the 2D BEV space and then calculate the 2D IoUs.
But for the \basename, we directly use the predicted 2D RoIs to compute the IoUs.
If the IoUs with all ground truth boxes are lower than $0.3$, the RoI is assigned to the background class.
Otherwise, it is assigned the box with the highest IoU.

Given the assignment of ground truth to the RoI, similarities between the ground truth and anchor poses are computed for non-background RoIs.
The anchor pose having the highest euclidean similarity is used as the classification target:
\begin{equation}
k_{target} = \arg\max_{k \in K} \sum_{j=1}^{J} || a_{k,j} - g_j ||,
\end{equation}
where $a_{k,j}$ is the position of joint $j$ of the $k$-th anchor pose, $g_j$ represents the joint $j$ of the ground truth, $J$ is the number of joints and $K$ is the number of anchor poses.
For computing the loss, we use a sparse cross entropy function given the target index $k_{target}$.




\subsubsection{2D Pose Refinement Loss}
For $L_{2D}$, we add the predicted 2D regression deltas to the anchor poses to obtain a set of final 2D predictions $P_{2D}$.
Using the IoU comparison method described above, we assign the target values $T_{2D}$ for each of the $N_{fg}$ foreground RoIs as the corresponding 2D ground truth poses.
The 2D regression loss is computed as a smooth L1 loss between the target poses $T_{2D}$ and the predicted pose proposals $P_{2D}$.
The regression loss is computed only for the foreground classes:
\begin{equation}
\label{eq:2d_loss}
L_{2D}(P_{2D}, T_{2D}) = \frac{1}{N_{fg}} \sum_{i=1}^{N_{fg}} l_{i} \cdot \texttt{smooth\_l1}(p_{i}, t_{i})
\end{equation}
where $l_{i}=1$ if $y_{i} > 0$ else $l_{i}=0$.

\subsubsection{3D Pose Refinement Loss}
For $L_{3D}$, we add the regressed deltas to the 3D anchor poses to obtain absolute 3D poses $P_{3D}$. Since the 3D ground truth is not available, we project the 3D poses into the 2D image space and compute the smooth L1 loss using a projection function $\texttt{Pr}(\cdot)$ and the 2D ground truth poses $T_{2D}$. Similar to the 2D loss, this is also computed for the foreground classes:
\begin{equation}
\label{eq:3d_loss}
L_{3D}(P_{3D}, T_{2D}) = \frac{1}{N_{fg}} \sum_{i=1}^{N_{fg}} l_{i} \cdot \texttt{smooth\_l1}(\texttt{Pr}(p_{i}), t_{i})
\end{equation}

\subsection{Implementation Details}
For \name, we trained our model for a total of 50 epochs with a batch size of 1, an Adam optimizer and an initial learning rate of $5e^{-5}$.
Learning rate is not decayed as the network is trained from scratch for both the inputs and so a higher value is required.
Whereas for the \basename, we trained our model for a total of 170 epochs with a batch size of 4 and an initial learning rate of $1e^{-3}$.
We decayed the learning rate by a factor of 0.8 after every 50 epochs and use a COCO pre-trained backbone.
RMSProp optimizer from the PyTorch library was used.
In order to make the networks direction-invariant, we extend the existing dataset with left-to-right flipped versions of the training set.
We flip the RGB image from left to right, followed by flipping the LiDAR point cloud along the x-axis.
Note that in our work, the x-axis represents the right direction and the origin lies at the camera center.
For the pose annotations, we represent the flipped x coordinate of the 2D pose in terms of the image width $w$ as $f(x) = w - x$.
Additionally, we filter out the samples having missing joints or missing segmented point clouds during the data loading phase.
For the post processing, we follow the pose proposals integration described in LCRNet~\cite{lcrnet}.

Overall, we introduced a novel architecture for multi-person 3D human pose estimation, using RGB and LiDAR data for in-the-wild scenarios of autonomous driving.

\section{Evaluation} \label{sec:evaluation}
We evaluated our \name~network architecture on the PedX~\cite{pedx} dataset and validated our \basename~against state-of-the-art on the MPII~\cite{mpii} dataset.
In contrast to MPII, the PedX dataset is new and has not yet been widely used.
The dataset has 9380 images with instance segmentation on pointclouds and 2D pose annotations.
3D bounding box annotations were generated by using the outer hull of the outlier cleaned 3D instance segmentation.
The dataset does not provide 3D pose annotations, which leads to our indirect performance evaluation via newly introduced metrics.
We use common evaluation metrics such as Percentage of Correct Keypoints (PCKh@0.5), 2D Mean Per Joint Position Error (MPJPE) and add new metrics for indirect 3D evaluation.
Center Point Depth Error (CDE) computes the axis aligned bounding box around the predicted pose and computes the depth error against the correct 3D bounding box.
Center Point X-Y Error (XYE) uses the same aligned bounding boxes and computes the error orthogonal to the depth, allowing separate inspection of error sources.
Therefore, these metrics can capture the absolute position error of the predictions.
\begin{table}[t]
\renewcommand{\arraystretch}{1.0}
\caption{Comparison of RGB Baseline vs \name~on PedX. LiDAR significantly improves the precision of 3D location (1/5 CDE, 1/3 XYE). 2D results improve slightly (MPJPE and PCKh@0.5).}
\label{tab:pedx_results}
\centering
 \begin{tabular}{|l|c|c|c|c|c|c|c|} 
 \hline
Model & Type & 2D MPJPE & PCKh & CDE & XYE\\[0.5ex]
\hline
RGB Base. [ours] & 2D &  87.76px & 65.02\% & - & -  \\
(RGB only) & 3D &  87.66px & 65.92\%  & 4.88m & 1.44m \\
\hline
\name~[ours]  & 2D & 45.66px & 70.08\% & - & -\\
(RGB + LiDAR) & 3D & \textbf{45.65px} & \textbf{70.22\%} & \textbf{0.95m} & \textbf{0.39m} \\
 \hline
\end{tabular}
\end{table}

Since there are no baselines on the PedX dataset, we implemented an \basename~(RGB only version of our model) similar to LCR-Net++\cite{lcrnet} and tested it on MPII~\cite{mpii} and PedX~\cite{pedx}.
Table~\ref{tab:verify_mpii_results} and Table~\ref{tab:verify_pedx_results} prove a similar performance to the original LCR-Net++ for our \basename.
The sole difference between the \basename~and \name~is in the LiDAR extension.
This allows us to attribute all performance gains over the baseline to adding LiDAR.

To show the improvements by including LiDAR, we compare our \basename~against our \name~with as identical parameters as possible.
Both networks were trained to optimal accuracy with similar parameters, the same training procedure and the same data.
The current state-of-the-art typically evaluates 3D performance root-joint relative.
With the availability of LiDAR, we can evaluate absolute 3D performance.
Most approaches only provide root relative results, however our \basename~and \name~produce absolute 3D predictions.
In our evaluation, we capture the error of the root joint by the CDE and XYE metrics introduced above.

\subsection{RGB Baseline vs \name}
\begin{table}[t]
\renewcommand{\arraystretch}{1.0}
\caption{RGB Baseline (inspired by LCRNet++) Verification on MPII}
\label{tab:verify_mpii_results}
\centering
 \begin{tabular}{|l|c|c|c|c|} 
 \hline
Model & Category & Type & 2D MPJPE & PCKh@0.5\\
\hline
LCRNet++\cite{lcrnet} & single & 2D & - & 74.61\% \\
RGB Baseline (ours) & single & 2D & \textbf{58.30px} & \textbf{81.95\%}\\
\hline
RGB Baseline (ours) & multi & 2D & 61.53px & 79.82\%\\
\hline
\end{tabular}
\end{table}
\begin{table}[t]
\renewcommand{\arraystretch}{1.0}
\caption{RGB Baseline (inspired by LCRNet++) Verification on PedX.}
\label{tab:verify_pedx_results}
\centering
 \begin{tabular}{|l|c|c|c|c|c|} 
 \hline
Model & Type & Trained On & 2D MPJPE & PCKh@0.5 \\
\hline
LCRNet++\cite{lcrnet} & 2D & non PedX  &  246.98px & 52.35\%  \\
LCRNet++\cite{lcrnet} & 3D & non PedX  &  250.60px & 47.44\%  \\
RGB Base. (ours) & 2D & non PedX  &  151.73px & 36.53\%  \\
\hline
RGB Base. (ours) & 2D & PedX  &  87.76px & 65.02\% \\
RGB Base. (ours) & 3D & PedX  &  \textbf{87.66px} & \textbf{65.92\%} \\
 \hline
\end{tabular}
\end{table}

The 2D MPJPE and PCKh@0.5 metrics capture improvements in the pose predictions.
Our \name~reduces the 2D MPJPE by a factor of $1.9$ and improves the PCKh for 2D and projected 3D by $+4.3\%$ (Table~\ref{tab:pedx_results}).
The improvements in CDE and XYE depict the performance of our model with respect to absolute positioning of the pose.
Here \name~reduces the CDE and XYE by a factor of $5.1$ and $3.7$ respectively (Table~\ref{tab:pedx_results}).
The best 3D object detectors specialized and evaluated on the very competitive KITTI~\cite{kitti} benchmark currently achieve errors of $0.11-0.22$m on pedestrians~\cite{fuerst2020lrpd}.
Our \name~significantly outperforms RGB only pose estimators and achieves 3D precision ($0.39$m XYE) almost similar to the state-of-the-art in pedestrian detection on KITTI.

Furthermore, we visually inspected the performance of our algorithm. Fig.~\ref{fig:occluded} shows a case where our \name~is able to precisely locate the pedestrian despite heavy occlusion by a silver SUV. In Fig.~\ref{fig:visuals}, we do a qualitative comparison of the \basename~and \name.
\begin{figure}[b]
    \begin{center}
	    \includegraphics[width=0.98\linewidth]{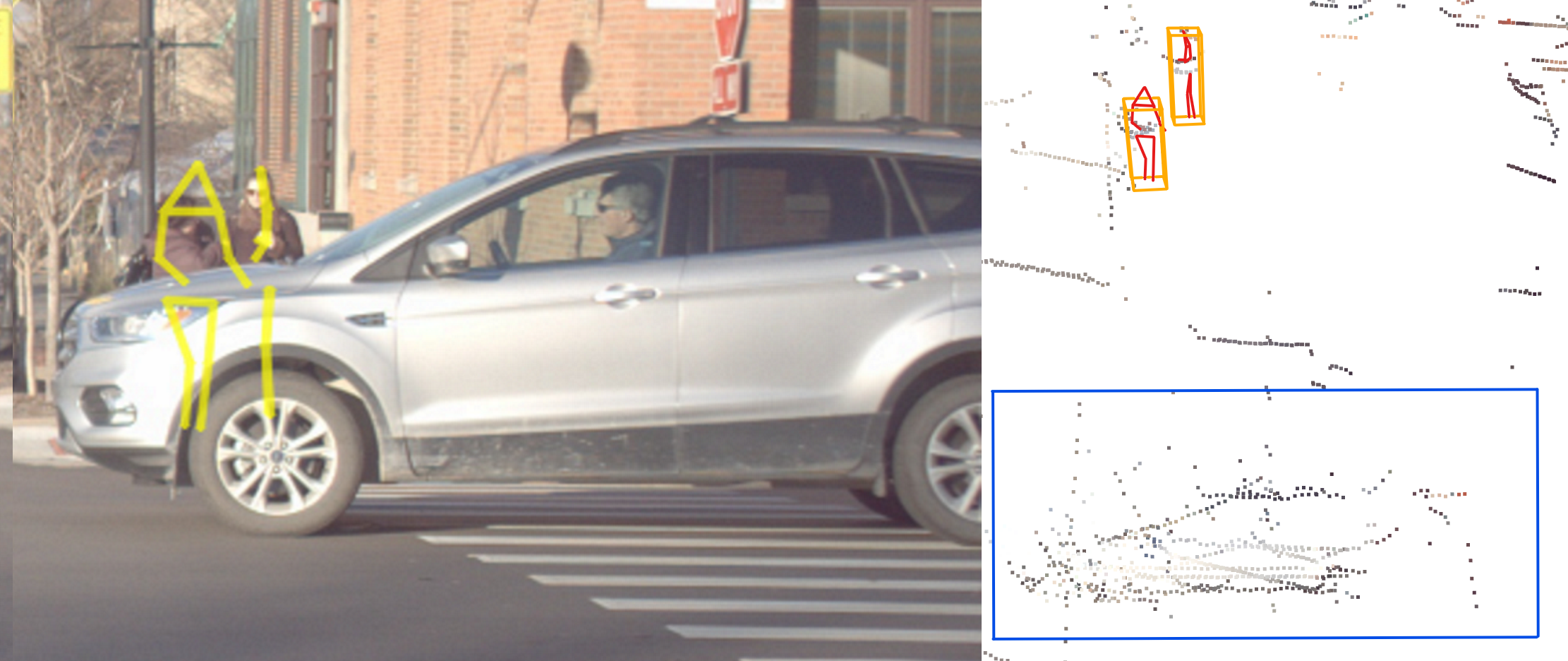}
    \end{center}
    \caption{Heavily occluded pedestrians can be located precisely with limited pose quality using the LiDAR pointcloud. The pedestrians occluded by the silver SUV (manually marked blue) are precisely located. 2D predictions are shown in yellow, 3D predictions in red, 3D ground truth in orange and the occluding car in blue.}
\label{fig:occluded}
\end{figure}
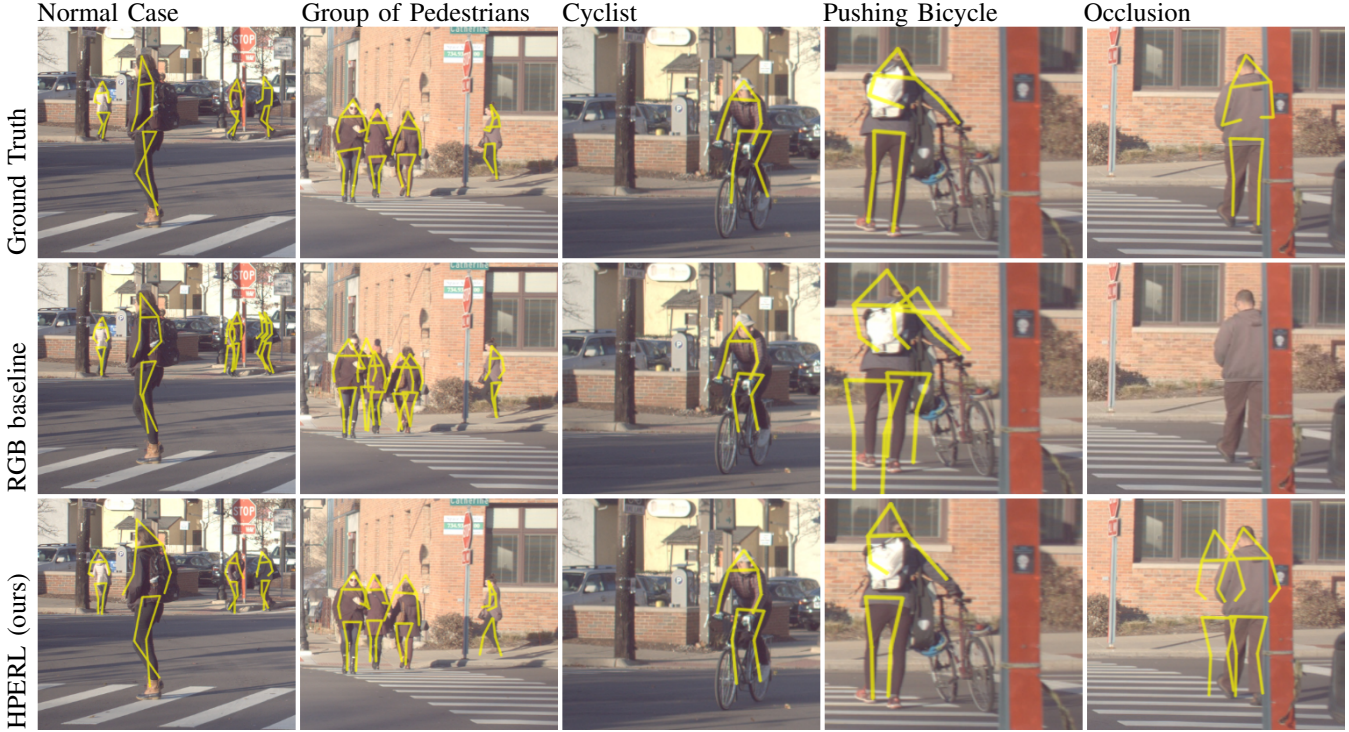
\begin{figure*}[t]
    \begin{center}
    \begin{overpic}[width=0.98\linewidth,,tics=10]{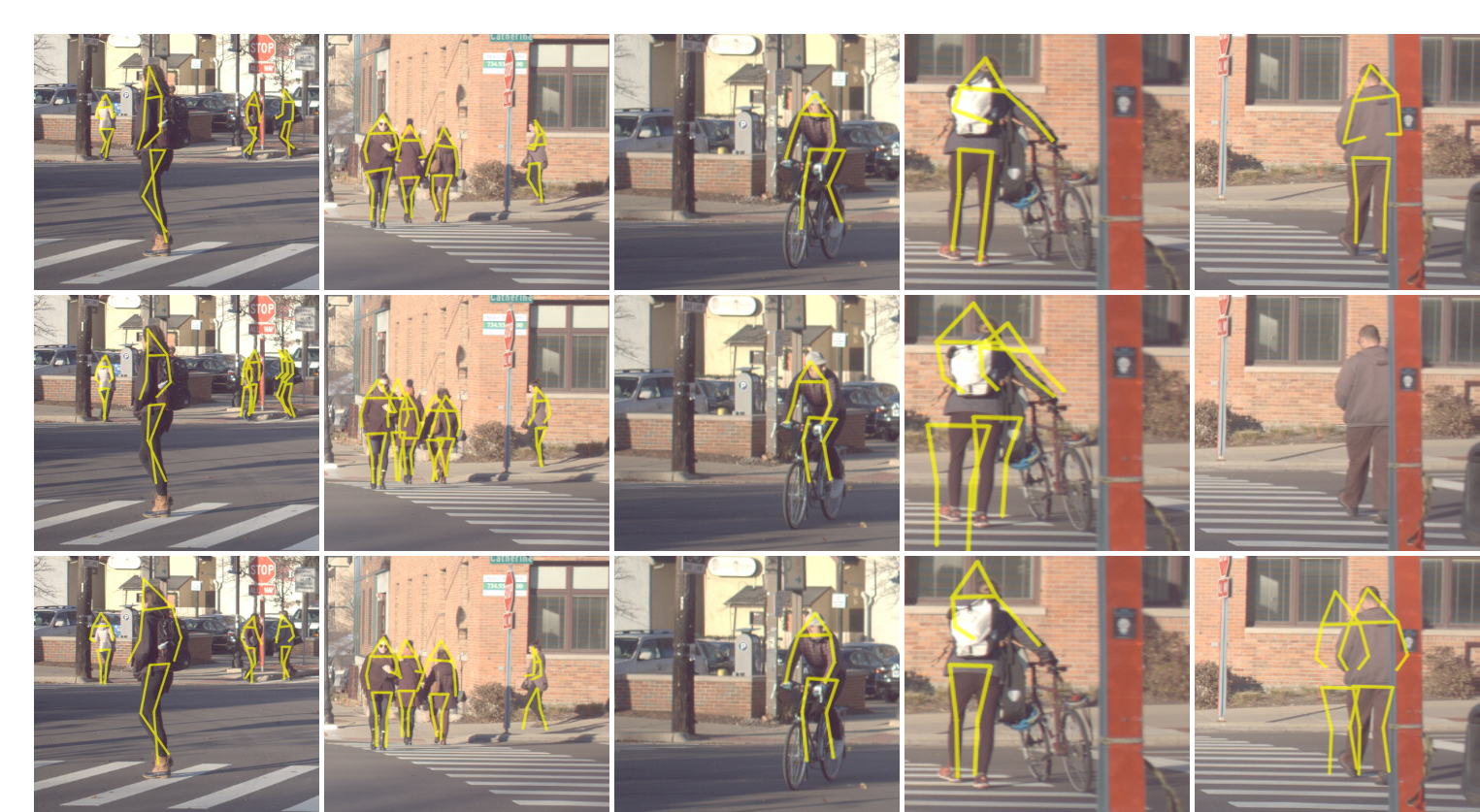}
	   \put(02.25,53.00){Normal Case}
	   \put(22.00,53.00){Group of Pedestrians}
	   \put(41.50,53.00){Cyclist}
	   \put(61.00,53.00){Pushing Bicycle}
	   \put(80.50,53.00){Occlusion}
	   \put(00.00,35.50){\rotatebox{90}{Ground Truth}}
	   \put(00.00,17.75){\rotatebox{90}{\basename}}
	   \put(00.00,00.00){\rotatebox{90}{\name~(ours)}}
	\end{overpic}
    \end{center}
    \caption{Qualitative comparison of performance between the \basename~and \name. The poses are depicted in yellow. In common scenarios shown on the left, both algorithms detect the pedestrians, but the baseline struggles with false positives at multiple depths. Albeit a rare case, the cyclist on the bicycle is well detected by both methods. Pushing a bicycle however causes false positives for \basename~and an imprecise detection for \name. Partial occlusions are difficult for both approaches, however \name~is able to detect the pedestrian but at the cost of a false positive.}
\label{fig:visuals}
\end{figure*}

\subsection{Ablation Studies}
To verify the effectiveness of all the components of our approach, we derived ablation studies. We changed the feature extractor, pre-training, internal network parameters and recorded the metrics (Table~\ref{tab:ablation}).
We observed that for the 3D performance (CDE and XYE), adding LiDAR information has the biggest performance impact.
Even poorly configured versions of \name~outperform the~\basename.

Analysing the results of the ablation study, we found that having a customized model with fewer parameters and less generalization gap outperforms initializing the model with Imagenet~\cite{imagenet} pre-trained weights.
For the fusion strategy, we observed that concatenation is better suited than the mean operation.
But for the data augmentation, we were able to see only a minor improvement, which is explained by the natural variance in poses and a roughly symmetrical distribution of poses regarding the LR-axis.
\begin{table*}[ht]
\renewcommand{\arraystretch}{1.0}
\caption{Ablation studies of HPERL on the PedX dataset. Feature extractor, pre-training, fusion, RoI operation and data augmentation were varied to determine the impacts on the 3D pose estimation on PedX. The biggest impact is due to adding LiDAR.}
\label{tab:ablation}
\centering
 \begin{tabular}{|l|c|c|c|c|c|c|c|c|c|c|c|c|} 
 \hline
Feature Extractor & Input & Pretrained & \#Features & Fusion & RoI Op. & Data Aug. & 2D MPJPE & PCKh@0.5 & CDE & XYE\\[0.5ex]
\hline
\hline
Resnet-50 & RGB & COCO & 256 & - & RoI Align & LR-Flip & 87.66px & 65.92\%  & 4.88m & 1.44m \\
\hline
\hline
VGG-16 & RGB + LiDAR & Imagenet & 512 & Concat & RoI Pool & No &  131.25px & 56.13\% & 2.38m & 1.10m\\
\hline
VGG-16 & RGB + LiDAR & Imagenet & 512 & Concat & RoI Align & No &  74.62px & 58.67\% & 1.50m & 0.64m \\
\hline
Resnet-50 & RGB + LiDAR & Imagenet & 1024 & Concat & RoI Align & No & 64.87px & 61.41\% & 1.27m & 0.56m\\
\hline
\hline
VGG-16 & RGB + LiDAR & No & 256 & Mean & RoI Pool & No &  80.39px & 63.07\% & 1.57m & 0.73m\\
\hline
VGG-16 & RGB + LiDAR & No & 256 & Concat & RoI Pool & No &  70.03px & 65.04\%  & 1.02m & 0.58m\\
\hline
VGG-16 & RGB + LiDAR & No & 256 & Concat & RoI Align & No &  60.28px & 65.87\%  & \textbf{0.85m} & 0.49m\\
\hline
VGG-16 & RGB + LiDAR & No & 256 & Concat & RoI Align & LR-Flip &  \textbf{59.52px} & \textbf{68.56\%}  & 0.99m & \textbf{0.49m}\\
\hline
\end{tabular}
\end{table*}
\section{Conclusions} \label{sec:conclusions}
In this paper, we presented HPERL using a fusion of RGB images and LiDAR point clouds to precisely locate pedestrians and predict their pose. 
This method was trained to detect the 3D human poses without using any 3D pose annotations.
Our approach applied an implicit formulation of the learning goal via projection and 3D bounding boxes to learn the 3D predictions.
Thus, we introduced the CDE and XYE metrics to capture the 3D precision of the predictions.
This opens up new opportunities to deploy human pose estimation in the wild.

Our research shows the versatility of a 3D detector's fusion schema.
In this work we used AVOD~\cite{avod} as a backbone, however all backbones following the two stage approach introduced by Faster R-CNN~\cite{fasterrcnn} are compatible with our proposed architecture.

The results of our empirical analysis demonstrate a promising performance, which can be attributed to the inclusion of LiDAR as an additional input modality. 
However, the lack of in-the-wild datasets hinders large scale evaluations and development.
We hope that our work encourages the creation of datasets and further research, enabling the usage of human pose estimation for autonomous vehicles and other applications requiring high absolute precision.


\section*{ACKNOWLEDGMENT}

The research leading to these results is funded by the German Federal Ministry for Economic Affairs and Energy within the project ”KI-Absicherung” (grant: 19A19005U).

\bibliography{references}
\bibliographystyle{ieeetr}

\end{document}